\pdfoutput=1
\documentclass[11pt]{article}
\usepackage[]{latex/acl}
\usepackage{tikz}
\newsavebox{\circleit}
\savebox{\circleit}{%
  \tikz[baseline=(char.base)]{%
    \node[shape=circle, draw, inner sep=0.5pt] (char) {it};%
  }%
}

\usepackage{times}
\usepackage{latexsym}
\usepackage{paralist}

\usepackage[T1]{fontenc}
\usepackage[utf8]{inputenc}

\usepackage{microtype}

\usepackage{algorithm}      % 提供 algorithm 环境
\usepackage{algpseudocode}  % 提供 algorithmic 环境，用于编写伪代码
\usepackage{amsfonts,amsmath,bm,xspace}
\usepackage{graphicx}
\usepackage{multirow}
\usepackage{booktabs}
\usepackage{makecell}
\usepackage{amsmath}
\usepackage{amssymb}
\usepackage{enumitem}
\usepackage{mathtools}
\usepackage{amsthm}
\usepackage[normalem]{ulem}
\usepackage{CJK}
\usepackage{soul}
\newcommand{\kaiti}[1]{\begin{CJK*}{UTF8}{gkai} #1 \end{CJK*}}
\soulregister{\kaiti}7
\newcommand{\method}{\textsc{Trans-Zero}\xspace}

\useunder{\uline}{\ul}{}
\graphicspath{ {./images/} }

\usepackage[compact]{titlesec}
\usepackage{sidecap}
\usepackage{tabularx}
\renewcommand{\paragraph}[1]{\noindent\textbf{#1}}
\title{ 
\method: Self-Play Incentivizes Large Language Models for Multilingual Translation Without Parallel Data
}
\author{Wei Zou$^{\spadesuit}$\footnotemark[2]\quad
Sen Yang$^{\spadesuit}$\footnotemark[2]\quad
Yu Bao$^{\heartsuit}$\quad Shujian Huang$^{\spadesuit}$\footnotemark[1]\quad Jiajun Chen$^{\spadesuit}$\quad Shanbo Cheng$^{\heartsuit}$\footnotemark[1] \\
  $^{\spadesuit}$National Key Laboratory for Novel Software Technology, Nanjing University \\
  $^{\heartsuit}$ByteDance Research \\
  \texttt{\{zouw,yangsen\}@smail.nju.edu.cn}, \texttt{\{huangsj, chenjj\}@nju.edu.cn}\\
  \texttt{\{baoyu.001,chengshanbo\}@bytedance.com}
}

\begin{document}
\maketitle
\renewcommand{\thefootnote}{\fnsymbol{footnote}}
\footnotetext[1]{Corresponding author}
\footnotetext[2]{Work done during internship at ByteDance Research}

\renewcommand{\thefootnote}{\arabic{footnote}}
\begin{abstract}
The rise of Large Language Models (LLMs) has reshaped machine translation (MT), but multilingual MT still relies heavily on parallel data for supervised fine-tuning (SFT), facing challenges like data scarcity for low-resource languages and catastrophic forgetting. 
To address these issues, we propose \method, a self-play framework that leverages only monolingual data and the intrinsic multilingual knowledge of LLM. 
\method combines a novel Monte-Carlo Tree Search, G-MCTS, with preference optimization, achieving strong translation performance that rivals supervised methods. 
Experiments demonstrate that this approach not only matches the performance of models trained on large-scale parallel data but also excels in non-English translation directions. 
Further analysis reveals that G-MCTS itself significantly enhances translation quality by exploring semantically consistent candidates through iterative translations, providing a robust foundation for the framework's success. 
Codes are available at \url{https://github.com/NJUNLP/trans0}

\end{abstract}

\section{Introduction}
The advent of Large Language Models~(LLMs) witnesses a fundamental shift in machine translation~(MT) paradigms from the supervised end-to-end training~\cite{vaswani2017transformer} to the sophisticated generation of fine-tuned language models~\cite{achiam2023gpt}.

Unlike various downstream tasks that gain impressive proficiency through lightweight instruction tuning, multilingual translations between languages still necessitate sufficient fine-tuning with parallel data for specific translation directions.
% annotation scarcity and SFT issue
%However, fine-tuning an LLM to translate between languages it mastered still necessitates substantial annotations in specific translation directions, 
Besides external human annotations, researchers like \citet{xu2024contrastive,li2024mt} obtain translation annotations or preferences from external LLM, as long as they prove multilingual capability.
Either way, it faces a notable deficit in data scarcity for less popular languages.
% external annotations
Moreover, issues arise as the multilingual translation fine-tuning scales up.
The reliance on one-on-one MLE supervision has been criticized for potential biases that clash with natural language's inherent multilingualism~\cite{zhu2024preference} and poses a risk of catastrophic forgetting.
Furthermore, exceeding multilingual annotations inversely dilutes the pre-trained knowledge in supported languages, thus degrading overall cross-lingual performance~\cite{xu2023paradigm,zhu2023multilingual}.
\citet{xu2024xalma} proposes a mixture-of-expert with hand-crafted route across language modules. 
However, the route and distributed overheads increase exponentially as the number of translation directions involved increases.

Whereas traditional fine-tuning scaling approaches a plateau, leveraging LLMs' inherent knowledge rather than external supervision for self-improvement is trending~\cite{chen2024sppo,kumar2024training}.
% \citet{kondo2024enhancing} proposes to stay aligned by excavating data from LM's pre-training phase for ongoing training, yet limits the post-training scaling.
% The self-play paradigm further advances in harnessing the LLM for self-improvement, providing potential beyond direct supervision, since the pre-trained knowledge serves infinite annotations.
% In this work, we introduce the self-play paradigm into the training of multilingual translation abilities and lead the LLMs to improve their translation ability by exploring their own translation space.
%We propose TransZero~(Trans0), a self-play approach that improves LLMs' translation ability by exploring their own translation space. Trans0 works with only monolingual inputs and requires no parallel data or external bilingual annotations.
However, adapting this approach to MT introduces two technical challenges.
% First, exploring cross-lingual inferences that cover the semantically aligned or better translation candidates beyond the LLM's initial cross-lingual alignments is non-trivial.
First, systematic \textit{cross-lingual exploration} requires navigating complex semantic spaces beyond simple prompt engineering.
Traditional LLM planning involves delicate prompt-based reasoning, even fine-tuning, which is generally unavailable for most scenarios.
Second, \textit{multilingual quality assessment} must overcome the limitations of data-dependent quality estimation~(QE) metrics and reward model training complexities.

In this work, we introduce \method.
This innovative self-play framework enables LLMs to bootstrap their multilingualism by strategically exploring their semantic space, achieving self-improvement for multilingual translation given only monolingual data.
We start by defining a Multilingual Translation Process~(MTP) that generates translations by interweaving the languages supported by LLM, so the inference is scaled up to explore more potential translations.
% For the first issue, 
First, we implement the Genetic Monte-Carlo tree search~(G-MCTS) upon MTP, exploring potential translations.
% For the second issue, 
Second, we harness the search to assess translation preferences based on intrinsic multilingual consistency.
Experiments verify that \method improves the lesser translations through iterative G-MCTS and preference optimizations given only monolingual data.
% Our contributions are threefold: 
We summarize our contributions as follows:
\vspace{-0.3mm}
\begin{itemize}
\setlength{\itemsep}{0.3em}     % 项目之间的垂直间距
\setlength{\parsep}{0em}      % 段落之间的距离
\setlength{\parskip}{0em}     % 段落前后的额外垂直空间
\setlength{\topsep}{0em}      % 列表前后的垂直空间
    \item First self-play framework extending to multilingual MT training with monolingual data.
    \item Novel integration of MCTS that explores improved translation for preference.
    \item An intrinsic translation preference without additional QE modules enables iterative self-improvement.
\end{itemize}

\section{Preliminary}
\subsection{Machine Translation via LLM}

Traditional machine translation depends on large parallel supervision in specific language pairs, which is limited by insufficient annotation in less popular translation directions.
The emergence of multilingual pre-trained language models has revolutionized the field, enabling impressive performance across various downstream tasks with minimal supervision~\cite{wei2021finetuned}. 
State-of-the-art (SOTA) LLM-based translation systems now achieve competitive results with significantly fewer annotations by leveraging supervised fine-tuning guided by sophisticated instructions~(see Appendix~\ref{6_sec:prompts} for details).
Notably, \citet{zhu2023multilingual} demonstrated that LLMs exhibit remarkable zero-shot and few-shot machine translation capabilities, even without explicit instruction formatting or exemplars. This highlights the intrinsic potential of LLMs for self-improvement beyond finely calibrated supervision, opening new avenues for resource-efficient translation paradigms.

\subsection{Monte-Carlo Tree Search}
Monte-Carlo Tree Search~\cite[MCTS,][]{browne2012survey,swiechowski2023monte} is a heuristic search algorithm proficient for complex decision processes such as the game of Go~\cite{silver2016alphago}.
MCTS proceeds through four steps: selection, expansion, simulation, and backpropagation.
% \vspace{-0.5em}
\begin{itemize}
\setlength{\itemsep}{0.3em}     % 项目之间的垂直间距
\setlength{\parsep}{0em}      % 段落之间的距离
\setlength{\parskip}{0em}     % 段落前后的额外垂直空间
\setlength{\topsep}{0em}      % 列表前后的垂直空间
    \item \textbf{Selection.} MCTS descends the tree from its root by selecting the top amongst child nodes by their upper confidence bounds~($\operatorname{UCB}$):
\begin{equation}
    \operatorname{UCB}(\alpha)=\nu(\alpha)+2\sqrt{\frac{\log N(A)}{1+N(\alpha)}},
    \label{6_eq:UCB}
\end{equation}
where node $\alpha$ is a child of node $A$, and $N(*)$ is the node's visit count, with $\nu(*)$ as its current utility. 
The utility of $\alpha$ is the cumulated rewards $Q(\alpha)$ averaged by its visit count: 
\begin{equation}
    \nu(\alpha)=\frac{Q(\alpha)}{N(\alpha)}
    \label{6_eq:utility}
\end{equation}
The $\operatorname{UCB}$ balances the exploration and exploitation in the heuristic search.

\item \textbf{Expansion.} Upon reaching a selected node, MCTS expands the tree by adding a new child node representing a possible move from the current state.
% MCTS adds a new child node for the selected node via possible moves from the current state.

\item \textbf{Simulation.} From the newly expanded node, a random simulation is performed until either the decision process concludes or reaches a maximum step limit, estimating the reward $r$ for this decision path.
% From the newly expanded node, a random simulation is performed to the end of the decision process or certain maximum steps, estimating the reward $r$ to be received if a decision were conducted.

\item \textbf{Backpropagation.} The obtained reward is propagated backward, updating the utility values of all nodes along the path from the expanded node to the root. 
This update refines the UCB values, progressively enhancing the efficiency of subsequent search iterations.
% The simulated reward tributes to utilities of nodes along the path from the expanded node back to the root. This process updates all relevant UCBs that incrementally improve further search.
\end{itemize}

\subsection{Self-Optimization in LLM}

Typically, optimizations for LLM rely on external rewards (e.g., critic and revise modules) for tuning~\cite{huang2022selfimprove,tian2024toward,zhang2024rest}. 
Recent work explores self-optimization using LLMs' internal knowledge, leveraging performance gaps across test scenarios. 
For example, in multilingual tasks, higher-performing languages can optimize lesser ones~\cite{geng2024whynot}. 
\citet{she2024mapo} uses a strong language as a pivot to map and optimize weaker languages. 
Self-play preference optimization~\cite[SPPO]{chen2024sppo} adopts the gaming theory where the post-update models shall prevail by a win rate $\mathbb{P}(\cdot)$ to optimize the preference. 
The SPPO is apt for pairwise preference $(y_w \succ y_l,x)$ in the following symmetric loss:
\begin{align}
    \mathcal{L}&_\text{SPPO}(x,y_w,y_l,\theta,\pi_\text{ref})\\ \nonumber
    =&[\log\frac{\pi_\theta(y_w|x)}{\pi_\text{pre}(y_w|x)}-\eta(\mathbb{P}(y_w\succ y_l|x)-\frac{1}{2})]^2 \\ \nonumber
    &-[\log\frac{\pi_\theta(y_l|x)}{\pi_\text{pre}(y_l|x)}-\eta (\mathbb{P}(y_l\succ y_w|x)-\frac{1}{2})]^2,
    \label{6_eq:SPPO_loss}
\end{align}
where $\theta$ is the LLM's parameters, $\pi_\text{pre}$ is the LLM before update and $\eta$ is a hyperparameter.
SPPO suits the human preferences' non-transitive, unstable nature, thus enabling sustainable self-play optimization.

\begin{figure*}[t]
\centering
\vspace{-5mm}
\includegraphics[width=\linewidth]{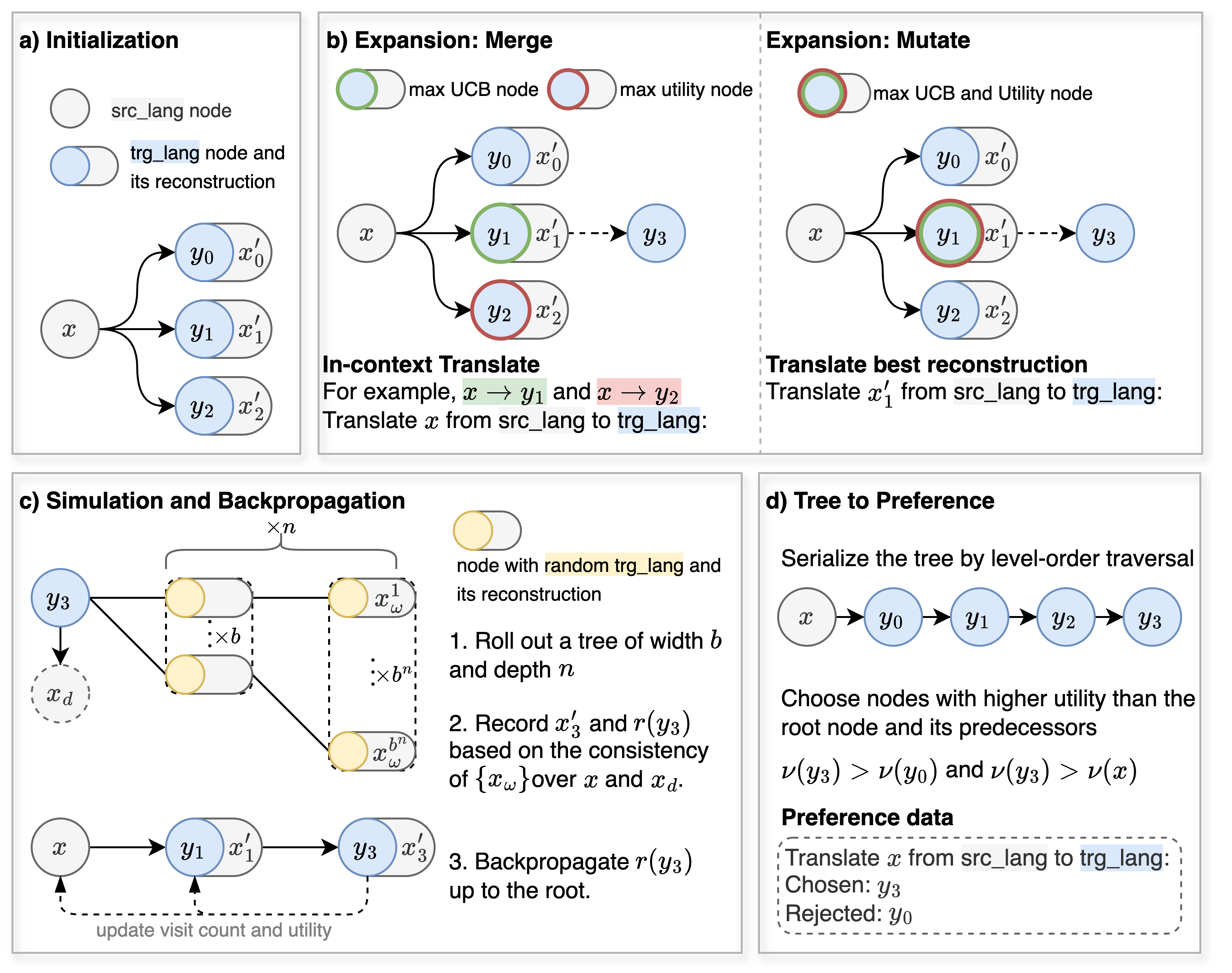}
\caption{
\textbf{Overview of \method.}
Once the tree is initiated, the search cycles the selection, expansion, simulation, and back-propagation for new nodes.
b) G-MCTS selects the node with maximum UCB to expand a new child node. 
There are two types of genetic expansion: merge and mutate.
c) A mass roll-out simulation of MTP trajectories assesses the semantic consistency.
The assessed reward is backpropagated to guide the search.
d) Finally, we harvest the search tree into data pairs for preference optimization.
}
\label{fig:overview}
\vspace{-2mm}
\end{figure*}

\section{\method}
In this section, we present \method. 
First, we introduce the Multilingual Translation Process~(MTP), a novel framework orchestrating multi-step translation across multiple languages~($\S$\ref{ss:mtp}). 
Second, we implement Genetic Monte Carlo Tree Search~(G-MCTS) upon MTP to explore promising translations~($\S$\ref{ss:g_mcts}), which derive preference from cross-lingual semantic consistency in the search purely through translation prompts, eliminating the need for explicit reasoning training or reward learning. 
Finally, we utilize the search results for further preference optimization~($\S$\ref{ss:tree_to_preference}), enabling the unsupervised MT training given only monolingual data.

\subsection{Multilingual Translation Process}\label{ss:mtp}
We define the Multilingual Translation Process~(MTP) as an iterative translation involving at least two languages, denoted as $\ \{L_i\}_{|\{L_i\}|>1}$, where each translation step maps the sentence from one language to another distinct language within the set.
An MTP trajectory of length $T$ starts from a source language $l$:
\begin{align*}
    l_T=f(l|l_1, l_2, ..., l_{T-1})  ,
\end{align*}
where the $l_i$ is a sentence in one language in $\{L_i\}$, and $f(\cdot)$ is the translation function.
MTP iteratively scales up the translation across languages, enabling similar cross-lingual preferences in work by \citet{geng2024whynot,she2024mapo}.

Notably, an optimized translation ensures that \textit{the semantics are maintained throughout the multilingual translations}.
Such consistency preference intuitively guides the search toward better translation candidates, which can also contribute to preference optimization.
E.g., translation optimized via round-trip translation promotes the bilingual semantic consistency given an MTP $x\rightarrow y\rightarrow x'$.

\subsection{Genetic Monte-Carlo Tree Search}\label{ss:g_mcts}
% The major difference 
As shown in Figure~\ref{fig:overview}, we conduct a genetic Monte-Carlo tree search based on the defined MTP.
The search employs genetic expansion coupled with semantic consistency simulation. 
Each tree node corresponds to a translation candidate in the target language and is generated autoregressively by translations from existing nodes\footnote{The translation instructions during G-MCTS are sampled from Table~\ref{tab:trans_prompt} in Appendix~\ref{6_sec:prompts}.}.

\paragraph{Initialization}
The input language is $X$ with input text $x$, and the output language is $Y$. 
We initialize the search tree with $x$ as the root and perform top-k sampling with a width of $b$ to generate $b$ translation candidates $\{y_i\}_b$ in language $Y$. 
Each candidate is assigned as a child node of the root. 
These nodes are quickly initialized by back-translation to language $X$, with the corresponding reconstruction recorded, denoted as $x^\prime$.
We define the consistency function $\operatorname{S}(x,x')$ over sentence pair $(x,x')$ of \textbf{same} language based on a mutual evaluation metric \( \operatorname{M}(x, x') \), e.g. BLEURT~\cite{sellam2020bleurt}. The consistency score is computed as:
\begin{equation}
    \operatorname{S}(x,x')=\frac{\operatorname{M}(x, x')+\operatorname{M}(x',x)}{2},
    \label{6_eq:consistency}
\end{equation}
where $x^\prime$ is to compute fast-initiated reward $r(y_i)=\operatorname{S}(x,x^\prime)$, followed by backpropagations.

\paragraph{Genetic Expansion}
Given an initialized search tree, each expansion step of the MCTS selects the node with the highest UCB value to generate a new translation in the target language.
However, a straightforward generation does not naturally lead to diverse exploration.
Inspired by genetic algorithms~\cite{sastry2005GA}, we propose two strategies for expansion based on the status of the current maximum UCB node:
% \begin{compactitem}
\begin{itemize}
\setlength{\itemsep}{0.3em}     % 项目之间的垂直间距
\setlength{\parsep}{0em}      % 段落之间的距离
\setlength{\parskip}{0em}     % 段落前后的额外垂直空间
\setlength{\topsep}{0em}      % 列表前后的垂直空间
    \item \textbf{Merge.} We perform a merging when the current maximum UCB node differs from the maximum utility node:
    Merging is a few-shot translation given the current best translation~(i.e., the maximum utility node) as demonstrations:
    \begin{align*}
        &y_t = f(x \mid y_{\text{UCB}}, y_{\nu}) \\
        &y_{\text{UCB}} = \arg\max_{y_{<t}} \{\operatorname{UCB}(y)\} \\
        &y_{\nu} = \arg\max_{y_{<t}} \{ \nu(y) \}
    \end{align*}
    Specifically, $(x, y_{{\nu}})$ and $(x, y_{\text{UCB}})$, the pairs from both the maximum utility node and the maximum UCB node, are prepended to the instruction. 
    The LLM then translates the \textbf{original input} $x$ to the target language, with the given context.
    \item \textbf{Mutate.} We perform a mutation when the current maximum UCB node is the same as the maximum utility node. 
    Mutation enables creative exploration based on existing translations, which is performed by translating a variant of the original input:
    \begin{align*}
        y_t = f(x' | \operatorname{arg}\max_{y_{<t}}\{\operatorname{UCB}(y)\}). 
    \end{align*}
    Specifically, the LLMs translate the \textbf{best} reconstruction $x^\prime\in\{x_\omega\}$ recorded during simulations of the parent node, instead of the original input $x$:

% \end{compactitem}
\end{itemize}

% The translation instructions within the above operations are sampled from Table~\ref{tab:trans_prompt} in Appendix~\ref{6_sec:prompts}.
% Each candidate is generated autoregressively given existing translations.
The merge operation reduces the most promising MTP trajectories via an in-context translation of the original $x$. 
Meanwhile, the mutation extends the existing trajectory through additional reconstructed $x^\prime$ from simulation, which may come from languages other than source and target~($\S$~\ref{ss:simulation} on simulations). 
Thus a search tree is constructed, with any path from the root to a node in the tree a valid MTP trajectory.

\begin{table*}[t]
\footnotesize
\vspace{-5mm}
\centering
% \small
\resizebox{\textwidth}{!}
{
\begin{tabular}{c|ll}
\toprule
\textbf{Utility~($\nu$)} & \multicolumn{2}{l}{\textbf{Tree Nodes}} \\
\midrule
0.5255 & \multicolumn{2}{l}{If you always harm others, the chickens gonna come home to roost.} \\
\midrule
0.5595 & \textcircled{1}\kaiti{如果你总是伤害别人，那么必然会有一天会有报应。} & (If you always hurt others, there will be retribution one day.) \\
1.2290 & \textcircled{2}\kaiti{你总是伤害别人，最后伤害的就是自己。} & (You always hurt others, and in the end you end up hurting yourself.) \\
0.5801 & \textcircled{3}\kaiti{你总是伤害别人，最后伤害的还是你自己。} & (You always hurt others, but in the end you end up hurting yourself.) \\
2.3230 & \textcircled{4}\kaiti{你总是伤害别人，最后总会有报应的。} & (You always hurt others, and you will get your comeuppance in the end.) \\
0.6117 & \textcircled{5}\kaiti{你若总是伤害别人，鸡就要飞回你的巢了。} & (If you keep hurting others, the chickens will fly back to your nest.) \\
0.5275 & \textcircled{6}\kaiti{如果你总是伤害别人，最后鸡蛋就要落回自己的头上。} & (If you always hurt others, eventually the eggs will fall back on your head.) \\
0.5626 & \textcircled{7}\kaiti{你总是伤害别人，最后鸡蛋就要落回自己的头上。} & (If you always hurt others, in the end the eggs will fall back on your head.) \\
0.5657 & \textcircled{8}\kaiti{如果你总是伤害别人，那么你也会遭到报应。} & (If you always hurt others, then you will also suffer retribution.) \\
0.5601 & \textcircled{9}\kaiti{如果你总是伤害别人，那么最后总会有报应的。} & (If you keep hurting others, you will get punished in the end.) \\
0.4676 & \textcircled{10}\kaiti{如果你总是伤害别人，那么麻雀总会飞回窝的。} & (If you always hurt others, the sparrow will always fly back to the nest.) \\
\bottomrule
\end{tabular}
}
\vspace{0.1cm} 
\resizebox{\textwidth}{!}
{
\begin{tabular}{l|l|ll}
\toprule
\multicolumn{4}{c}{\textbf{Extracted Preference Pairs for Self-Play Preference Optimization~(SPPO)}} \\
% \hline
\midrule
\textbf{Chosen} & \textbf{Rejected} & \multicolumn{2}{c}{\textbf{Win rates (softmax)}} \\
% \toprule
\midrule
\textcircled{2}\kaiti{你总是伤害别人，最后伤害的就是自己。} & \textcircled{1}\kaiti{如果你总是伤害别人，那么必然会有一天会有报应。} & 1.23 : 0.56 & (0.6614) \\
\textcircled{4}\kaiti{你总是伤害别人，最后总会有报应的。} & \textcircled{2}\kaiti{你总是伤害别人，最后伤害的就是自己。} & 2.32 : 1.23 & (0.7491) \\
\textcircled{4}\kaiti{你总是伤害别人，最后总会有报应的。} & \textcircled{3}\kaiti{你总是伤害别人,最后伤害的还是你自己。} & 2.32 : 0.58 & (0.8511) \\
\textcircled{7}\kaiti{你总是伤害别人，最后鸡蛋就要落回自己的头上。} & \textcircled{6}\kaiti{如果你总是伤害别人，最后鸡蛋就要落回自己的头上。} & 0.56 : 0.52 & (0.5088) \\
\textcircled{8}\kaiti{如果你总是伤害别人，那么你也会遭到报应。} & \textcircled{7}\kaiti{你总是伤害别人，最后鸡蛋就要落回自己的头上。} & 0.57 : 0.56 & (0.5008)\\
\bottomrule
\end{tabular}

}
\caption{\textbf{Example of Tree-to-Preference.} We perform level-order traversal of a search tree for English-to-Chinese translation. 
The first line presents the source input as the root. 
The Chinese translations are accompanied by corresponding English explanations enclosed in parentheses.
Given that the utility near the root shall be larger, we apply a sorting algorithm to arrange the sequence in descending order, where each swap during the sort makes a preference pair.
% \textbf{Note:} Duplicate nodes are reduced to the ancestor closest to the root, with utilities accumulated during traversal. 
% The win rate is calculated using the softmax function: $\text{win rate} = \frac{\exp(\nu_i)}{\exp(\nu_i) + \exp(\nu_j)}$.
}
% \caption{
% Level-order traversal (from the root node to the leaf nodes) of a search tree for English-to-Chinese translation. 
% The first line presents the source input as the root. 
% The Chinese translations are accompanied by corresponding English explanations enclosed in parentheses.
% Given that the utility values near the root shall be larger, we apply a sorting algorithm to arrange the sequence in descending order, where each swap operation during the sort makes a preference pair.
% }
\label{tab:6_trie2seq}
\end{table*}

\begin{figure}[t]
    \small
    \centering
    \includegraphics[width=0.8\linewidth]{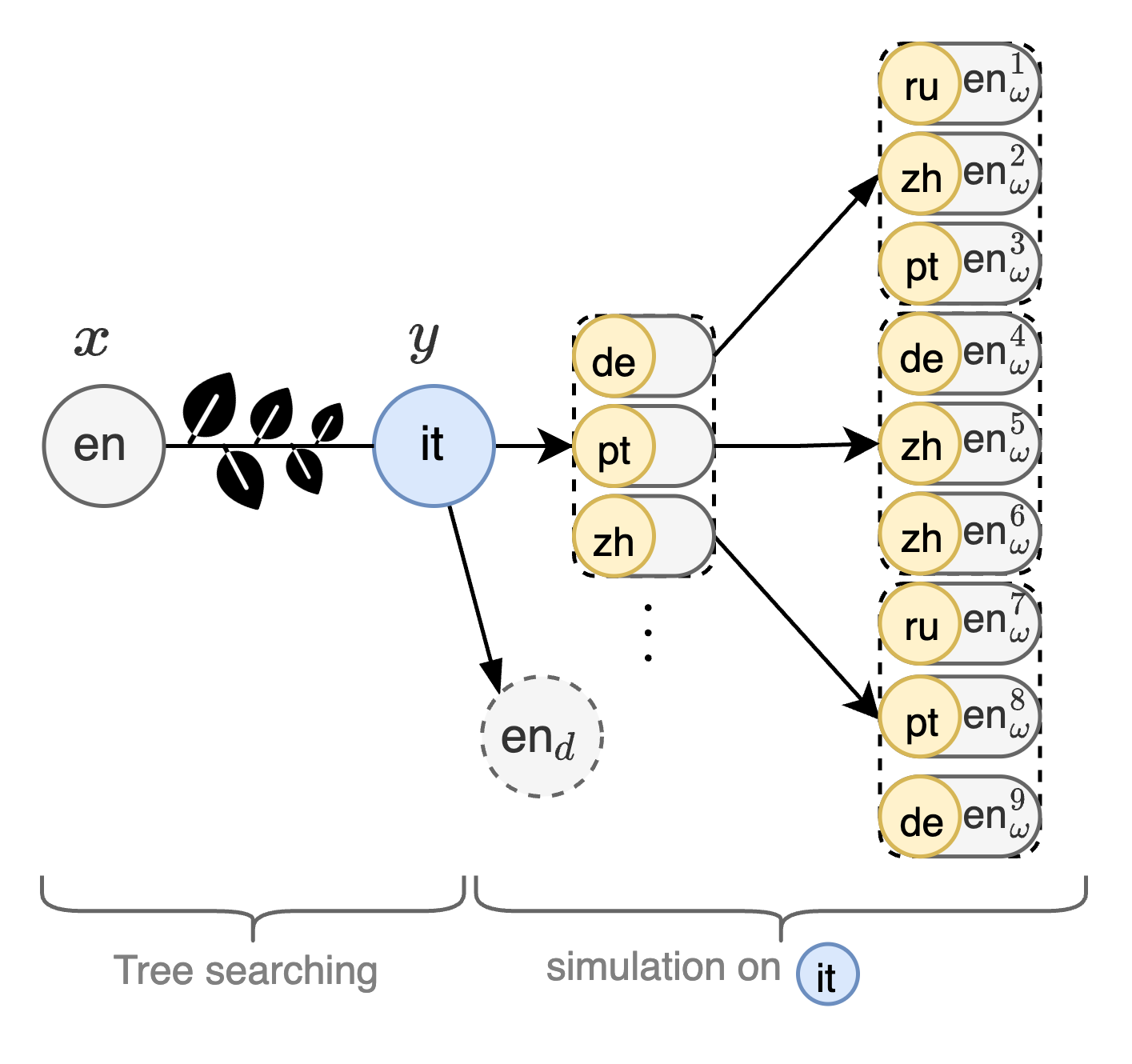}
    \caption{
    \textbf{An example of simulation on English-to-Italian translation candidate using $b=3$ and $n=2$.} Through roll-outs of MTP, the Italian candidate~(\usebox{\circleit}) is assessed by semantics consistency of $b^n$ English reconstructions $\{\text{en}_\omega^1, \cdots, \text{en}_\omega^9\}$ from simulated trajectories.
    }
    \label{fig:simulate}
\end{figure}

\paragraph{Simulation with Multilingual Semantic Consistency}\label{ss:simulation}
The simulation generates reward signals through semantic agreement across translations, which directly updates the UCB values in tree search to assess translation preferences without external supervision.
The key insight is that a better translation candidate achieves higher semantic consistency for themselves and their descendants, thus guiding an optimized search.

As shown in Figure~\ref{fig:simulate}, assessing candidate $y$ involves rolling out a temporary sub-tree on $y$ through MTP with a width of $b$ until it reaches a maximum depth of $n$. 
Each rollout step translates the parent node into a different language sampled from the supported languages $\{L_i\}$, with corresponding input reconstruction $x_\omega$. 
Ultimately, this results in $b^n$ MTP trajectories with reconstructions $\{x_\omega\}_{b^n}$. 
Given Eq.~\ref{6_eq:consistency}, the semantic consistency of these reconstructions $\{x_\omega\}_{b^n}$ with the original input $x$ can be calculated as follows:
\begin{align*}
    r(y)= \max(\underbrace{\overline{\operatorname{S}(x_\omega,x)}}_{\text{literal}},\underbrace{\overline{\operatorname{S}(x_\omega,x_d)}}_{\text{free}}), {x_\omega \in \{x_\omega\}_{b^n}}
\end{align*}
where $x$ is the original input, and $x_d$ is a direct reconstruction of the candidate $y$.
Translation can be literal or free, with free translations assessed via straightforward back-translation $x_d$.
Each assessment is \textit{averaged} over all trajectories' $\{x_\omega \}_{b^n}$, with the superior one as the reward $r(y)$ derived from the simulation.
The best simulation is recorded as $x^\prime$ for further expansions and simulations:
\begin{align*}
    x^\prime=\arg\max_{\{x_\omega\}_{b^n}}(S(x_\omega,x), S(x_\omega, x_d))
\end{align*}

During the backpropagation, the reward $r(y)$ updates the utility $\nu$ and visit counts of all nodes on the trajectory from the current node back to the root according to Eq.~\ref{6_eq:utility}.

\begin{algorithm}[t]
\caption{Tree-to-Preference Algorithm}
\label{alg:preference_extraction}
\begin{algorithmic}[1]
\Require Translation candidates $\{y_*\}$ with utilities $\{\nu_*\}$, search tree $\mathcal{T}$ rooted from $x$
\Ensure Preference pairs with win rates for SPPO

\State \textbf{Step 1: Serialize and Sort Tree}
\State $\mathcal{S} \gets \text{LevelOrderTraversal}(\mathcal{T})$ \Comment{Serialize tree and merge duplicates}
\State $\mathcal{S} \gets \text{SelectionSort}(\mathcal{S}, \text{descending})$ \Comment{Sort nodes by utility}

\State \textbf{Step 2: Generate Preference Pairs}
\State $\mathcal{P} \gets \emptyset$
\For{each swap $(y_i, y_j)$ in $\mathcal{S}$}
    \If{$\nu_i > \nu_{\text{root}}$ \textbf{and} $\nu_i > \nu_j$}
        \State \text{win rate:}$\mathbb{P} \gets \frac{\exp(\nu_i)}{\exp(\nu_i) + \exp(\nu_j)}$
        \State $\mathcal{P} \gets \mathcal{P} \cup \{(y_i \succ y_j, \mathbb{P}|x)\}$
    \EndIf
\EndFor

\State \textbf{Return} $\mathcal{P}$ \Comment{Preference pairs with win rates for SPPO}
\end{algorithmic}
\end{algorithm}

\subsection{Tree-to-Preference Algorithm}\label{ss:tree_to_preference}
As Table~\ref{tab:6_trie2seq} shows, once the G-MCTS is finished, we extract preference data from the translation candidates $\{y_*\}$ based on their utility $\{\nu_*\}$ by Algorithm~\ref{alg:preference_extraction}. 
The tree is serialized where duplicate nodes are merged to their ancestor, with utilities and visit counts accumulated.
Intuitively, nodes away from the root take more MTP steps to generate and thus face more risk of semantic loss as the translation step increases. 
Consequently, if a node has a higher utility than its ancestors or siblings, the corresponding translation is preferred for optimization.
Furthermore, the root node tracks a comprehensive utility, which reflects the expected semantic consistency of the entire search. 
Thus, the utility of the preferred node shall also be higher than that of the root node.
The preference pairs are utilized in SPPO, where their utilities are transformed into win rates through a softmax function.
Note that translations that failed language detection may be generated during MTP, and their utility is halved as a penalty during the sorting and filtering.

\begin{table*}[t]
\small
\centering
\tabcolsep 5.5pt
\begin{tabular}{lccccccccccc}
\toprule
                        & \multicolumn{2}{c}{\textbf{EN$\Rightarrow$X}} & \multicolumn{1}{c}{} & \multicolumn{2}{c}{\textbf{X$\Rightarrow$EN}} & \multicolumn{1}{c}{} & \multicolumn{2}{c}{\textbf{X$\Rightarrow$X}} &  & \multicolumn{2}{c}{\textbf{Average}} \\
\cmidrule{2-3} \cmidrule{5-6} \cmidrule{8-9} \cmidrule{11-12} 
& \textbf{BLEURT} & \textbf{KIWI} &  & \textbf{BLEURT} & \textbf{KIWI} &  & \textbf{BLEURT} & \textbf{KIWI} &  & \textbf{BLEURT} & \textbf{KIWI} \\ 
\midrule
Mixtral-8x7B-Instruct   & 55.42 & 69.07 &  & 75.41 & 81.63 &  & 54.49 & 71.64 &  & 61.77 & 74.11 \\
Llama3.1-Instruct       & 62.57 & 74.28 &  & 72.07 & 77.90 &  & 62.52 & 76.49 &  & 65.72 & 76.22 \\
Qwen2.5-Instruct        & \underline{72.16} & 81.99 &  & 77.70 & 84.60 &  & \underline{68.59} & 79.87 &  & \underline{72.82} & 82.15 \\
ALMA                    & 71.98 & 82.60 &  & \underline{78.25} & 84.34 &  & 61.07 & 80.89 &  & 70.43 & 82.61 \\
ALMA-R                  & 69.38 & \underline{83.10} &  & 77.52 & \underline{84.87} &  & 51.03 & \underline{82.47} &  & 65.98 & \underline{83.48} \\
Tower-Instruct          & \textbf{76.74} & \textbf{85.26} &  & \textbf{78.73} & \textbf{85.02} &  & \textbf{72.98} & \textbf{83.08} &  & \textbf{76.15} & \textbf{84.45} \\
\midrule
Llama3.1-Base    & 33.18 & 25.53 &  & 50.83 & 55.64 &  & 34.38 & 53.52 && 39.46 & 44.89 \\
\quad w/ SFT (40k)   & \underline{74.46} & 82.30 &  & 77.30 & 84.03 &  & 71.23 & 80.02 &  & 74.33 & 82.12 \\
\quad w/ SFT (5m) & \textbf{75.80} & \textbf{84.61} &  & \textbf{78.47} & \textbf{84.61} &  & \textbf{73.30} & \underline{82.33} && \textbf{75.86} & \textbf{83.85} \\
\quad w/ \textbf{\method}   & 73.71 & \underline{83.20} &  & \underline{77.60} & \underline{84.34} &  & \underline{73.28} & \textbf{82.71} && \underline{74.86} & \underline{83.42} \\
\midrule
Qwen2.5-Base     & 62.91 & 73.98 &  & 70.98 & 80.50 &  & 62.70 & 77.86 && 65.53 & 77.45 \\
\quad w/ SFT (5m)  & \textbf{75.32} & \textbf{84.79} &  & \textbf{78.21} & \textbf{85.42} &  & \textbf{72.99} & \textbf{82.92} && \textbf{75.49} & \textbf{84.38} \\
\quad w/ \textbf{\method}       & \underline{75.05} & \underline{84.48} &  & \textbf{78.21} & \underline{84.88} &  & \underline{72.23} & \underline{82.30} && \underline{75.16} & \underline{83.89} \\
\bottomrule
\end{tabular}
    \caption{
    \textbf{\method achieves comparable and improved translation compared to SFT baselines with only monolingual self-play.} We highlight the best and the second-best performances in each section in \textbf{bold} and \underline{underlined}, respectively.}
    \label{tab:main_results}
\end{table*}

\section{Experiments}
% 1. What we want to verify: inference-scaling improves lesser translation results without training.
% 2. The preference fine-tuning improves the model.
% 3. Fully zero-shot translation agent.

\subsection{Settings}
% We extract the initial instructions for inference from Flores-200~\cite{costa2022no}, with only one sample from each translation direction.
\paragraph{Data.}
We conduct experiments across six widely used languages: English (EN), German (DE), Portuguese (PT), Italian (IT), Chinese (ZH), and Russian (RU).
We utilize the latest monolingual data from the WMT datasets. 
The SFT data for baselines is generated from the combination of the Flores-200 development set, with equal size for all translation directions.
To evaluate multilingual translation performance, we employ the Flores-200 benchmark~\cite{nllb2022flores}, assessing three key translation directions: (1) EN$\Rightarrow$X~(English to other languages), (2) X$\Rightarrow$EN~(other languages to English), and (3) X$\Rightarrow$X~(inter-translations between non-English languages).
These evaluations cover all translation directions of the six languages mentioned above.

\paragraph{Metrics.}
We evaluate translation quality with the reference-oriented metric BLEURT~\cite{sellam2020bleurt} and reference-free metric COMET-KIWI~\cite[KIWI,][]{rei-etal-2022-cometkiwi}.

\paragraph{Baselines.}
We compare \method with the following representative baselines:
\begin{compactitem}
    \item \textbf{General Instruct LLMs}: LLMs with off-the-shelf instruct-following ability for MT, e.g., Mixtral-8$\times$7B, Llama3.1-8B-instruct and Qwen2.5-7B-instruct.
    \item \textbf{MT-oriented LLMs}: LLMs supervised by MT annotations, e.g., ALMA~\cite{xu2023paradigm} with parallel annotations, ALMA-R~\cite{xu2024contrastive}
    with preference annotations and Tower-Instruct~\cite{colombo2024tower} with multi-task MT-related annotations, representing strong supervised baselines.   
    \item \textbf{Base model and SFT model}: The base LLM for \method, and their supervised counterparts.

\end{compactitem}

\paragraph{Implementation.}
The training is conducted on 32 NVIDIA A100 GPUs (80GB). 
We utilize two base LLMs without instruction tuning: Llama-3.1-8b and Qwen-2.5-7b. 
Since some base LLMs~(e.g., Llama-3.1-Base) lack the initial instruction-following for translation, we cold-start the LLM with a snippet of translation instructions (Appendix~\ref{6_sec:implement_detail}). 
The \method is parallelized across 32 threads, each assigned a batch of 10 sentences for random translation directions for G-MCTS. 
Upon completion of the search, we reduce all search threads for filtered preference data and apply Self-Play Preference Optimization~\cite[SPPO]{chen2024sppo}.
Additional details are in Appendix~\ref{6_sec:implement_detail}.

\subsection{Main Results}

% scaling inference as a direct improvement
% training on preferences harvest from the 
As shown in Table~\ref{tab:main_results}, \method based on Llama3.1 and Qwen2.5 achieves performance comparable to MT baselines trained on large-scale annotations, despite using only monolingual data for self-play. 
While \method matches the English translation performance (EN$\Rightarrow$X) of ALMA-R, which also utilizes preference optimization, it significantly surpasses ALMA-R in non-English translation directions. 
Compared to Tower-instruct, an LLM trained on large-scale annotations, \method exhibits slightly lower performance but remains highly competitive.

We also evaluated the translation performance of the base model and its instruction fine-tuned version for comparison. 
Through exploration learning, \method significantly enhances the performance of both base models.

Additionally, we include SFT baselines using $5m$ parallel data by pairing the Flores200 development set, and cold-start instructions ($5k$) as parallel data. 
Although \method does not match the performance by $5m$ supervision on EN$\Rightarrow$X and X$\Rightarrow$EN, it shows significant improvements in the non-English translation direction (X$\Rightarrow$X), achieving performance comparable to $5m$ supervision.

\begin{figure}[t]
    \centering
    \includegraphics[width=0.9\linewidth]{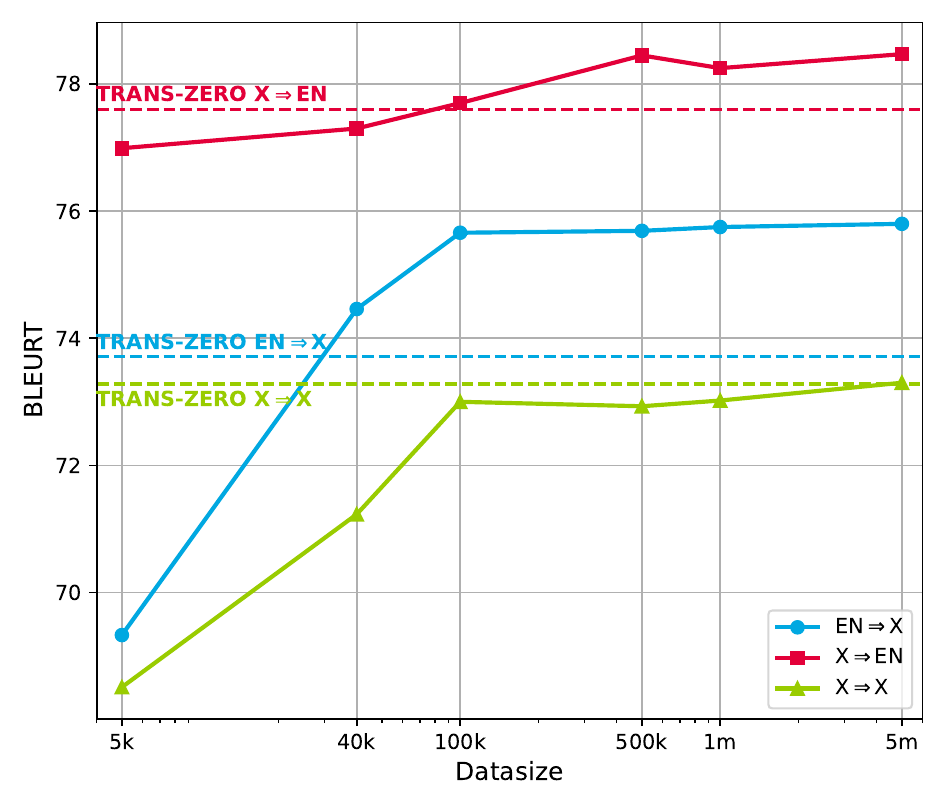}
    \caption{
    \textbf{BLEURT performance for SFT based on the Llama3.1-Base at different data sizes.} We include the performance of \method in each language direction.
    }
    \label{fig:abl_sft_data}
\end{figure}

We further compared the performance improvement to varying scales of parallel annotations in SFT. 
As shown in Figure~\ref{fig:abl_sft_data}, the translation quality saturates after more than 100k samples, especially for the EN$\Rightarrow$X and X$\Rightarrow$X directions.
This suggests that simply increasing the amount of parallel annotations may not lead to proportional translation improvements.

\begin{figure}[t]
    \centering
    \includegraphics[width=0.9\linewidth]{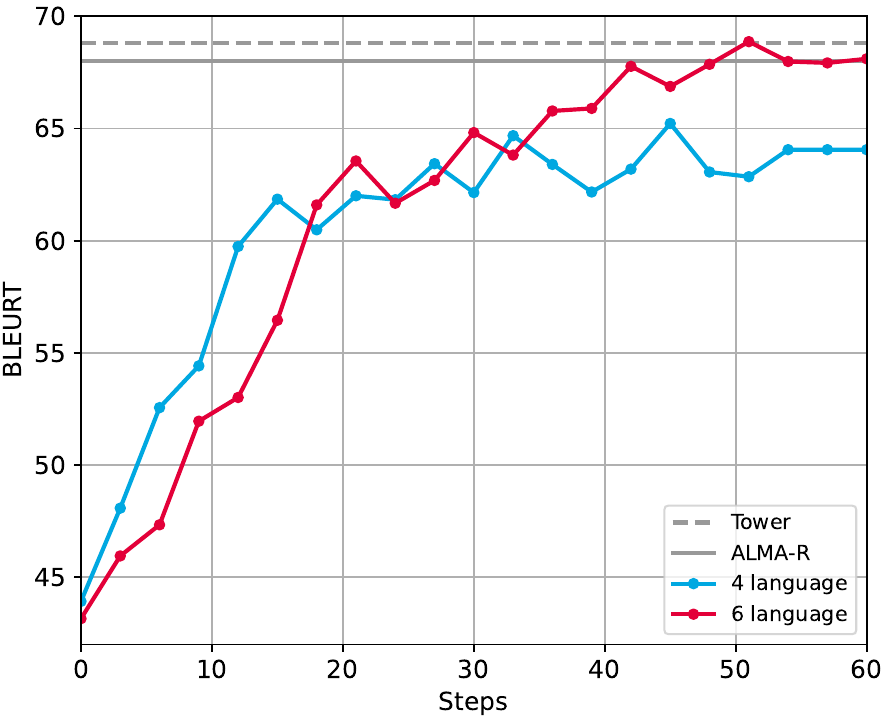}
    \caption{
    \textbf{The learning diagram of \method on Llama3.1-Base for German-to-Chinese translation demonstrates the search process in 4-language and 6-language settings under G-MCTS.}
    By incorporating 6 languages, \method attains BLEURT scores on par with the baseline systems.
    }
    \label{fig:self_improve}
\end{figure}

Increasing the number of languages expands the potential exploration thus improving the \method's performance upper bound. 
We explored with $4$ and $6$ languages for the G-MCTS. 
Figure~\ref{fig:self_improve} illustrates the performance changes in German-Chinese translation during Llama3.1-8b training. 
The number of languages used in tree search significantly impacts \method's performance upper bound: increasing the number of languages can enhance the overall learning performance. 
With 6 languages, \method essentially matches the performance of the open-source baseline system.

\begin{table*}[t]
\small
\vspace{-5mm}
\centering
\tabcolsep 4pt
\begin{tabular}{lllcllcllcll}
\toprule
& \multicolumn{2}{c}{\textbf{EN$\Rightarrow$X}} & \multicolumn{1}{c}{} & \multicolumn{2}{c}{\textbf{X$\Rightarrow$EN}} & \multicolumn{1}{c}{} & \multicolumn{2}{c}{\textbf{X$\Rightarrow$X}} &  & \multicolumn{2}{c}{\textbf{Average}} \\
\cmidrule{2-3} \cmidrule{5-6} \cmidrule{8-9} \cmidrule{11-12} 
& \textbf{BLEURT} & \textbf{KIWI} &  & \textbf{BLEURT} & \textbf{KIWI} &  & \textbf{BLEURT} & \textbf{KIWI} &  & \textbf{BLEURT} & \textbf{KIWI} \\ 
\midrule
ALMA-R 
& 69.38 & 83.10 &  & 77.52 & 84.87 &  & 51.03 & 82.47 &  & 65.98 & 83.48\\
\quad + G-MCTS 
& Failed & Failed &  & Failed & Failed &  & Failed & Failed &  & Failed & Failed\\ 
\midrule
Tower-Instruct 
& 76.74 & 85.26 &  & 78.73 & 85.02 &  & 72.98 & 83.08 &  & 76.15 & 84.45\\
\quad + G-MCTS 
& 76.44$_\text{-0.30}$ & 85.33$_\text{+0.07}$ &  & 78.28$_\text{-0.45}$ & 85.12$_\text{+0.10}$ &  & \textbf{74.42}$_\textbf{\text{+1.44}}$ & 83.57$_\text{+0.49}$ &  & 76.38$_\text{+0.23}$ & 84.67$_\text{+0.22}$\\ 
\midrule
Llama3.1-Base 
& 33.18 & 25.53 &  & 50.83 & 55.64 &  & 34.38 & 53.52 &  & 39.46 & 44.89\\
\quad + G-MCTS 
& Failed & Failed &  & Failed & Failed &  & Failed & Failed &  & Failed & Failed\\ 
\midrule
Llama3.1-Instruct 
& 62.57 & 74.28 &  & 72.07 & 77.90 &  & 62.52 & 76.49 &  & 65.72 & 76.22\\
\quad + G-MCTS 
& \textbf{64.21}$_\textbf{\text{+1.64}}$ & \textbf{80.12}$_\textbf{\text{+5.84}}$ &  & 70.01$_\text{-2.06}$ & \textbf{79.86}$_\textbf{\text{+1.96}}$ &  & \textbf{68.12}$_\textbf{\text{+5.60}}$ & 77.12$_\text{+0.63}$ &  & \textbf{67.45}$_\textbf{\text{+1.73}}$ & \textbf{79.03}$_\textbf{\text{+2.81}}$\\ 
\midrule
Llama3.1-SFT (5k) 
& 69.33 & 80.19 &  & 76.99 & 83.97 &  & 68.51 & 78.38 &  & 71.61 & 80.85\\
\quad + G-MCTS 
& \textbf{71.55}$_\textbf{\text{+2.22}}$ & \textbf{82.23}$_\textbf{\text{+2.04}}$ &  & 76.89$_\text{-0.10}$ & 84.00$_\text{+0.03}$ &  & \textbf{71.92}$_\textbf{\text{+3.41}}$ & \textbf{81.23}$_\textbf{\text{+2.85}}$ &  & \textbf{73.45}$_\textbf{\text{+1.84}}$ & \textbf{82.49}$_\textbf{\text{+1.64}}$\\ 
\midrule
Llama3.1-SFT (5m) 
& 75.80 & 84.61 &  & 78.47 & 84.61 &  & 73.30 & 82.33 &  & 75.86 & 83.85\\
\quad + G-MCTS 
& 76.16$_\text{+0.36}$ & 84.95$_\text{+0.34}$ &  & 78.71$_\text{+0.24}$ & 84.68$_\text{+0.07}$ &  & 73.36$_\text{+0.06}$ & 82.48$_\text{+0.15}$ &  & 76.08$_\text{+0.22}$ & 84.04$_\text{+0.19}$\\ 

\bottomrule
\end{tabular}
\caption{
\textbf{G-MCTS enhances translation by scaling up inference, given the models' own instruction-following capability and multilingualism.}
Performance improvements beyond one point are highlighted in \textbf{bold}.
The base LLM and ALMA-R exhibit limitations due to their failure to follow instructions in various translation directions.
The search particularly enhances X$\Rightarrow$X translations, where the availability of SFT annotations is significantly limited compared to English-related annotations.
}
\label{tab:infer_scaled}
\vspace{-2mm}
\end{table*}

% \subsection{G-MCTS-Based Inference Enhancement}
\subsection{Inference-time Scaling with G-MCTS}
We further investigate the G-MCTS's potential to exploit high-quality translations as an inference-time scaling.
Inference time scaling, such as Chain of Thought~(CoT)~\cite{wei2022cot}, has become popular for improving the performance of LLMs. 
Notably, traditional CoT requires additional learning given multiple natural language understanding supervisions. 
In contrast, G-MCTS enables straightforward inference-time scaling using only translation instructions (computation overheads detailed in the Appendix~\ref{6_sec:overhead}). 
During the tree search, merging and mutation continuously explore and integrate relevant expressions from the multilingual semantic space, revising translations based on the LLM's conditional generation capabilities.

We employ G-MCTS with 6 languages to explore translations for the LLama-3.1 baselines, as well as Tower-Instruct and ALMA-R. 
The candidate of the highest utility makes the final translation.
As Table~\ref{tab:infer_scaled} shows, G-MCTS requires a language model with basic instruction-following capabilities. 
Consequently, the Llama3.1-Base model fails the search due to its lack of instruction-following in various translation directions. 
Similarly, ALMA-R also fails due to its limited multilingualism, as indicated by its significantly lower X$\Rightarrow$X performance.

In contrast, Tower-Instruct and Llama3.1-Instruct significantly improve translation performance in the X$\Rightarrow$X direction, benefiting from multilingual priors. 
The base model trained with small-scale supervision also shows notable improvements.
However, the improvement upon Llama3.1-SFT~(5m) with large-scale supervision, is almost negligible.
This suggests that when translations are fully activated, the performance gains from G-MCTS, rooted in LLM's inherent multilingualism, are not statistically significant.

\section{Related Work}
% LLM for MT
The utilization of LLM for machine translation has become popular, aligning with the prevailing trends in LLM applications.
\citet{xu2023paradigm} first shifts the machine translation paradigm to fine-tuned LLM with moderate parallel supervision.
Though LLM seems more data-efficient, it does not have a big appetite for large-scale supervision due to potential catastrophic forgetting~\cite{xu2023paradigm,kondo2024enhancing}. 
Directly scaling up multilingual MLE supervision hurts performance on resource-rich languages~\cite{xu2024xalma}.
Therefore, XALMA~\cite{xu2024xalma} hand-craft a mixture-of-expert to route different translation directions through separated modules, while ALMA-R~\cite{xu2024contrastive} turn to scale up more expensive preference tuning.

Preference tuning offers flexibility when fitting LLM with subjective human expectations for open-ended generations. 
However, the expense of preference annotation has led researchers to seek more cost-effective data sources, e.g., with additional assessments such as critic and revise modules~\cite{huang2022selfimprove,tian2024toward,zhang2024rest}
Intuitively, the cross-lingual gaps in LLM offer a more scalable self-improvement preference as the proficient languages improve the lesser ones~\cite{geng2024whynot}, e.g., a straightforward improvement roots in the direct mapping from the dominant linguistic ability as preference~\cite{she2024mapo}.
Researchers further scale the preference by iterative and competitive gaming theory~\cite[SPPO]{chen2024sppo}, making it possible for models to self-improve.
Recent work by Deepseek~\cite{guo2025deepseek} has empirically validated its self-improving potential by employing large-scale reinforcement learning with its multilingual reasoning abilities.

\section{Conclusion}
In this work, we present \method, a novel framework for multilingual machine translation that leverages multilingual LLMs with monolingual data only. 
Our experiments demonstrate that the proposed Genetic Monte-Carlo Tree Search (G-MCTS) effectively enhances translation quality by exploiting the LLM's inherent multilingual and instruction-following capabilities. 
Furthermore, we show that iterative training of G-MCTS, combined with preference optimization using monolingual data, \method achieves scalable performance improvements, with the number of supported languages positively correlating with final translation quality. 
These findings establish a new direction for resource-efficient MT by shifting the paradigm from supervised parallel data to self-supervised monolingual learning.

\section*{Limitations}
% \method requires at least $600$ times more inference computations during a search for one input than baselines.
% \method requires more inference computations when searching for one input than baselines.
% Its training relies solely on the LLM's inherent multilingualism, making it incompatible with weaker models.
% Constrained by computational resources, we are unable to conduct more extensive self-play with larger LLMs or explore a broader range of languages during the search process.
% To further elevate the upper bound of \method, it is also crucial to explore which types of monolingual data are most cost-effective for learning.

While \method demonstrates promising results, it has several limitations that warrant discussion.
First, the framework introduces higher computational overhead than supervised baselines, as the search process requires extensive exploration of the cross-lingual semantic space. 
Second, its effectiveness is inherently tied to the multilingual capabilities of the underlying LLM, rendering it less suitable for weaker models with limited cross-lingual alignment. 
Third, due to computational constraints, our experiments were limited in scale: we were unable to explore larger LLMs or extend the search process to a broader range of languages. 
Finally, the framework's performance upper bound may be influenced by the quality and diversity of monolingual data used for training, highlighting the need for future research into identifying the most cost-effective data types for self-supervised training.

\section*{Ethics Statement}
The authors declare no competing interests. The datasets used in the training and evaluation come from publicly available sources and do not contain sensitive content such as personal information.

\section*{Acknowledgement}
We would like to thank the anonymous reviewers for their insightful comments. 
Shujian Huang and Shanbo Cheng are the corresponding authors. 
This work is supported by the National Science Foundation of China (No. 62376116, 62176120), research project of Nanjing University-China Mobile Joint Institute, and the Fundamental Research Funds for the Central Universities (No. 2024300507).

% \section*{Acknowledgements}

% Acknowledgment

% Entries for the entire Anthology, followed by custom entries
\bibliography{official}

\newpage
\appendix

\section{Translation Prompts}\label{6_sec:prompts}
\method adopts all the mainstream translation instructions as prompts in Table~\ref{tab:trans_prompt}.
The G-MCTS adopts random instructions during sampling, and preference optimization adopts random instructions for the extracted preference pairs.
LLMs for MT follow their default instructions for validation if available.
\method and other instruct-LLM baselines without default instruction adopt the ALMA instruction and <LABAL> as generation prompt for SFT and validation. 
All LLMs follow their default chat templates and generation prompts if available.

The translation context also samples instructions from Table~\ref{tab:trans_prompt} to organize translation pairs, which are then prepended to the translation instructions for in-context generation.

\begin{table}[ht]
\centering
\small
\resizebox{0.5\textwidth}{!}{
\begin{tabular}{l|l}
\toprule
Model          & Translation Instruction                                                    \\ 
\toprule
ALMA \&        & Translate this from \{src\_lan\} to \{trg\_lan\}:                          \\
ALMA-R         & \textbackslash n\{src\_lan\}: \{src\_sent\} \textbackslash n\{trg\_lan\}:     
                                        \\ 
\midrule
Tower-Instruct & Translate the following text from \{src\_lan\}           \\
               & into \{trg\_lan\}.\textbackslash n\{src\_lan\}: \{src\_sent\}  \\
               & \textbackslash n\{trg\_lan\}: \\
\midrule
Others         & Please translate the \{src\_lan\} into \{trg\_lan\}:         \\ 
               & \{src\_sent\} \\
\cline{2-2} 
               & \{src\_lan\}: \{src\_sent\} = \{trg\_lan\}:                                \\ 
\cline{2-2} 
               & \{src\_sent\} in \{src\_lan\} can be translated        \\ 
               & to \{trg\_lan\} as:  \\
\cline{2-2} 
               & \{src\_lan\}: \{src\_sent\} \textbackslash n \{trg\_lan\}:  \\ 
\cline{2-2} 
               & Explain the following \{src\_lan\} sentence  \\
               & in \{trg\_lan\}: \{src\_sent\} \\
\bottomrule
\end{tabular}
}
\caption{Commonly adopted translation instructions for LLM, \{src\_lan\} and \{src\_lan\} indicates the corresponding languages for source and target, and \{src\_sent\} presents the input sentence of the source language.  }
\label{tab:trans_prompt}
\end{table}

\section{Implementation Details}\label{6_sec:implement_detail}
For SFT baselines, we employ full parameter fine-tuning with a batch size of $1024$. 

For Llama3.1 cold-start, we generate approximately $5k$ translation instructions by \textbf{one random sentence tuple} from the Flores-200 development set, each representing one translation direction organized by a random instruction in Appendix~\ref{6_sec:prompts}. 
The cold-start applies 1-epoch LoRA fine-tuning with rank \( r = 64 \) and scaling parameter \( \text{Lora}_\alpha = 128 \), optimized by AdamW at learning rate \( lr = 1e-4 \).

For the G-MCTS, we configure a search width of \( b = 5 \) and a simulation depth of \( n = 2 \).
The tree is fast-initiated with \( b = 5 \) child nodes on the root by sampling, with a maximum of 20 additional tree nodes per search.
Note that the expansion in G-MCTS involves 6 times the sentence length during merging, while short sentences are not apt for diverse exploration, so we limit the G-MCTS to the sentence length within [30,256].

During a single self-play session, we distribute a monolingual batch across all devices, each launching a tree search for one sentence at a time, targeting a random translation direction.
The search results are gathered from all devices, with preference pairs extracted and organized by sampled translation instructions~(Table~\ref{tab:trans_prompt}) for one SPPO epoch.

The SPPO follows all the parameter settings from DPO, including the coefficient $\eta=\frac{1}{\beta}=10$, where $\beta$ is the temperature parameter of DPO.
The SPPO learning rate shall depend on the data size, where we adopt  \(lr = 1e-6\) given approximately $25k$ preference pairs gathered, with the batch size of $10k$ preference pairs.
One shall reduce the learning rate if more preference pairs are gathered during one self-play training epoch. 
Since preference extraction results may vary significantly on different sentences, we recommend scaling up the self-play session on a large sentence batch to ensure a stable training data flow.

The inference-time scaling with G-MCTS searches by the width of \( b = 10 \) given $6$ languages (English, German, Portuguese, Italian, Chinese, and Russian) with simulation depth of \(n=2\).

\section{Computational Overhead of G-MCTS} \label{6_sec:overhead}

The G-MCTS imposes substantial computational demands through its augmented inference architecture. 
Node-level operations exhibit significant latency differentials: merge and mutation operations require $6\times$ and $4\times$ prolonged in-context inference, respectively. 
Simulation for each node dynamically generates a subtree with \(b^n\) nodes. 
The self-play paradigm introduces multiplicative inference loads, where each monolingual training sample necessitates \(20 \times 5^2 = 500\) times more inference, while the inference-time scaling in our experiments introduces \(20 \times 10^2 = 2000\) times more inference.

% The training is on $32$ NVIDIA A100 GPUs (80GB).
% The baseline systems for supervised training~(SFT) use full parameter fine-tuning with a batch size of $1024$.
% We adopt two base LLM without instruction training: Llama-3.1-8b and Qwen-2.5-7b. 
% Since the some base models~(e.g., Llama3.1-Base) can not initially follow translation instructions, we generate approximately $5,000$ translation instructions from the flores200 development set, each presenting a translation direction.
% Each instruction is sampled from Table~\ref{tab:trans_prompt}.
% This data is used to cold-start the LLMs' instruction-following.
% \method adopts 1-epoch LoRA ($r=64, \text{Lora}_\alpha=128$) on instructions for cold-start.

% G-MCTS searches by width of $b=5$ with simulation depth of $n=2$, and a maximum of $20$ tree nodes.
% The search is parallel distributed across $32$ threads, with each thread processing a batch of 
% $10$ sentences assigned to random translation directions.
% Upon completion of all searches, we apply SPPO~\cite{chen2024sppo} optimization to the filtered preference data, with batch size $1024$.

% Further implementation details can be found in Appendix~TODO.

% This is an appendix.

\end{document}